%% file: main.tex
\documentclass[runningheads]{llncs}

\usepackage{eccv}

\usepackage{eccvabbrv}

\usepackage{graphicx}
\usepackage{booktabs}
\usepackage{hyperref}

\usepackage[accsupp]{axessibility}  

\usepackage{hyperref}

\usepackage{orcidlink}

\usepackage[misc]{ifsym}
\usepackage{tabu}
\usepackage{color,xcolor}
\usepackage{colortbl}

\usepackage{algorithm}
\usepackage{algpseudocode}
\usepackage[linesnumbered,algo2e]{algorithm2e}
\newcommand{\hrulealg}[0]{\vspace{1mm} \hrule \vspace{1mm}}

\definecolor{mygray}{gray}{.85}
\definecolor{mygray1}{gray}{.7}
\definecolor{mygray2}{gray}{.93}

\makeatletter
\newcommand{\thickhline}{%
    \noalign {\ifnum 0=`}\fi \hrule height 1pt
    \futurelet \reserved@a \@xhline
}

\newcommand{\teammembers}[1]{\noindent\textit{\textbf{Members:}} #1 \par}
\newcommand{\teammember}[1]{\noindent\textit{\textbf{Member:}} #1 \par}
\newcommand{\teamaffiliations}[1]{\noindent\textit{\textbf{Affiliations:}} {\parindent0pt \par #1} \par \vspace{2mm}}
\newcommand{\teamaffiliation}[1]{\noindent\textit{\textbf{Affiliation:}} {#1} \par \vspace{2mm}}

\newcommand{\nth}{\textsuperscript{th} }

\begin{document}

\title{LSVOS Challenge Report: Large-scale Complex and Long Video Object Segmentation} 

\titlerunning{LSVOS Challenge Report}

\author{Henghui Ding\textsuperscript{*~\Letter}\orcidlink{0000-0003-4868-6526}\and Lingyi Hong\textsuperscript{*}\and Chang Liu\textsuperscript{*}\and Ning Xu\textsuperscript{*}\and \\ Linjie Yang\textsuperscript{*}\and Yuchen Fan\textsuperscript{*}\and 
Deshui Miao\and Yameng Gu\and Xin Li\and Zhenyu He\and Yaowei Wang\and Ming-Hsuan Yang\and 
Jinming Chai\and Qin Ma\and Junpei Zhang\and Licheng Jiao\and Fang Liu\and 
Xinyu Liu\and Jing Zhang\and Kexin Zhang\and Xu Liu\and LingLing Li\and
Hao Fang\and Feiyu Pan\and Xiankai Lu\and  Wei Zhang\and Runmin Cong\and  
Tuyen Tran\and 
Bin Cao\and Yisi Zhang\and Hanyi Wang\and Xingjian He\and Jing Liu
}

\authorrunning{H.~Ding et al.}

\institute{}

\maketitle

\vspace{-5mm}
\renewcommand{\thefootnote}{\fnsymbol{footnote}}
\footnotetext[1]{ECCV 2024 LSVOS Workshop \& Challenge organizers. All others are challenge participants from the top teams of VOS and RVOS tracks.}
\footnotetext[0]{${\textrm{\Letter}}$ henghui.ding@gmail.com, Institute of Big Data, Fudan University}

\input{sections/0_abstract}

\input{sections/1_intro}

\input{sections/3_challenge}

\bibliographystyle{splncs04}
\bibliography{main}
\end{document}

%% file: sections/0_abstract.tex
\begin{abstract}
  Despite the promising performance of current video segmentation models on existing benchmarks, these models still struggle with complex scenes.  In this paper, we introduce the 6th Large-scale Video Object Segmentation (LSVOS) challenge in conjunction with ECCV \ECCVyear{} workshop. This year's challenge includes two tasks: Video Object Segmentation (VOS) and Referring Video Object Segmentation (RVOS). In this year, we replace the classic YouTube-VOS and YouTube-RVOS benchmark with latest datasets MOSE, LVOS, and MeViS to assess VOS under more challenging complex environments. This year's challenge attracted 129 registered teams from more than 20 institutes across over 8 countries. This report include the challenge and dataset introduction, and the methods used by top 7 teams in two tracks. More details can be found in our \href{https://lsvos.github.io/}{homepage}.
  \keywords{Large-scale Video Object Segmentation \and LSVOS \and Video Segmentation \and Complex Scenes \and MOSE \and MeViS \and LVOS}
\end{abstract}

%% file: sections/1_intro.tex
\section{Introduction}
\label{sec:intro}

Video object segmentation (VOS)~\cite{ding2024pvuw,xu2018youtube,perazzi2016benchmark,KhoRohrSch_ACCV2018,li2023transformer,wu2024towards,yang2019video}, is a fundamental problem in computer vision, focusing on tracking and segmenting target objects across video frames. Over the past few years, a lot of datasets and challenges are proposed.
Among them, YouTube-VOS~\cite{xu2018youtube} marks the first large-scale dataset. With its extensive collection of video sequences and annotations, it has facilitated the development of more robust and scalable VOS models. Based on the dataset, the Large-scale Video Object Segmentation (LSVOS) challenges is introduced. Since 2018, the challenge has been held for five consecutive years annually, and has become one of the most influential benchmarks. Meanwhile, with a large number of participants from around the world, LSVOS is also a crucial platform for showcasing advancements and addressing emerging issues in the field of VOS.

As VOS models are achieving notable success on existing benchmarks and past year's challenges~\cite{xu2018youtube,perazzi2016benchmark}, it seems that the task of VOS has already been well addressed. However, in contrast, some recent studies~\cite{MeViS,ding2023mose,hong2023lvos,hong2024lvos,GREC} also suggests that current models still face significant challenges when applied to realistic and complex scenes. These findings raises a question: how well are the performance of existing VOS models in \textit{real scenarios}? Thus, we shift our focus towards more challenging and realistic benchmarks, and introduce the 6\nth Large-scale Video Object Segmentation (LSVOS) challenge. This year's challenge includes two tasks: Video Object Segmentation (VOS) and Referring Video Object Segmentation (RVOS). Featuring with three latest and more challenging datasets, MOSE~\cite{ding2023mose}, LVOS~\cite{hong2023lvos,hong2024lvos}, and MeViS~\cite{MeViS}, we replace the classic YouTube-VOS~\cite{xu2018youtube} and Refer-Youtube-VOS~\cite{seo2020urvos} benchmark, to evaluate VOS under more challenging condition and real-world scenarios.

The 6\nth LSVOS challenge attracted significant international participation, with 129 teams from more than 20 institutes across over 8 countries. The competition culminated in 6 top-performing solutions. The collective efforts and achievements of this year's LSVOS challenge not only brought forward novel methodologies but also set the stage for future developments in video understanding.

%% file: sections/3_challenge.tex
\section{The 6\nth LSVOS Challenge}

\subsection{Track 1: Video Object Segmentation}
The Video object segmentation (VOS) task aims to segment a specific object instance throughout an entire video sequence given only the object mask of the first frame~\cite{DBLP:conf/cvpr/PerazziKBSS17,jang2017online,DBLP:conf/cvpr/JampaniGG17,xiao2018monet,ISFP,hu2018motion,han2018reinforcement,youtube_vos,cheng2018fast,xu2019mhp,chen2020state,huang2020fast,wug2018fast,lin2019agss,zhang2019fast}. This year, we replace the origin YouTube-VOS with MOSE~\cite{ding2023mose} and LVOS~\cite{hong2023lvos,hong2024lvos}. MOSE dataset includes 2,149 videos with annotations for 5,200 objects, encompassing a total of 431,725 segmentation masks. A key feature of MOSE is its focus on scenes with heavy crowding and occlusion, where target objects are frequently obstructed or disappear from view. LVOS consists of 720 sequences with an average duration of approximately 1.14 minutes, which is significantly longer than previous benchmarks. The final testing data for the task is randomly sampled from the test sets of both MOSE and LVOS datasets, which presents significantly increased difficulty with emphasis real-world complex and dense scenes, and impose higher requirements on VOS models, particularly in terms of maintaining accurate temporal associations and re-detecting objects.
\vspace{-3mm}

\subsection{Track 2: Referring Video Object Segmentation}
Referring video object segmentation (RVOS) aims to segment objects in video sequences based on language expressions~\cite{seo2020urvos, liu2024primitivenet,liu2023gres,liu2023multi, DsHmp,wang2019asymmetric,phraseclick,ye2021referring,ding2021vision,VLT}. Traditionally, language captions in current RVOS datasets have focused on salient objects and static attributes, often neglecting the dynamic aspect of motion across video frames~\cite{MeViS,seo2020urvos}. From this point, we replace the Refer-Youtube-VOS dataset that is used in past challenges with the latest motion-expression based referring VOS dataset, MeViS~\cite{MeViS}. Utilizing motion descriptions to refer to target objects imposes higher demands on the model's temporal understanding ability. The MeViS dataset addresses this gap by incorporating motion-based references. MeViS consists of 2,006 videos, with annotations for 8,171 objects, encompassing over 443,000 segmentation masks and 28,570 expressions. This extensive dataset significantly surpasses existing language-guided video segmentation datasets in terms of annotation scale and complexity. For the final testing phase of the RVOS task, the test data is randomly sampled from the test set of MeViS dataset.

\subsection{Evaluation Metrics}
Following previous works~\cite{MeViS,ding2023mose,hong2023lvos,hong2024lvos,perazzi2016benchmark,xu2018youtube}, both tracks utilize the commonly recognized metrics: Jaccard ($\mathcal{J}$) and F-measure ($\mathcal{F}$). The Jaccard $\mathcal{J}$ index measures the overlap between the predicted and ground truth regions, while the F-measure $\mathcal{F}$ assesses the precision and recall of contour detection. The average of the two scores ($\mathcal{J} \& \mathcal{F}$) is used as the overall performance metric. The final ranking of methods is determined by this average, calculated on the test set.

\begin{table}[t]
    \renewcommand\arraystretch{1.2}
    \centering
    \caption{\textbf{VOS Track results and final rankings.}}
    \vspace{-3mm}
    \setlength{\tabcolsep}{3.96mm}{\begin{tabular}{clrrr}
            \thickhline
             \rowcolor{gray!60} Rank & Team  & \multicolumn{1}{l}{$\mathcal{J}$} & \multicolumn{1}{l}{$\mathcal{F}$} & \multicolumn{1}{l}{$\mathcal{J\&F}$} \\
            \hline
            1                    & yahooo         & 80.90 & 76.16 & 85.63 \\
            \rowcolor{gray!20}2  & yuanjie        & 80.84 & 76.42 & 85.26 \\
            3                    & Xy-unu         & 79.52 & 75.16 & 83.88 \\
            \rowcolor{gray!20}4  & MVP-TIME       & 75.79 & 71.25 & 80.33 \\
            5                    & bai\_kai\_shui & 75.77 & 71.22 & 80.31 \\
            \rowcolor{gray!20}6  & NanMu          & 75.75 & 71.26 & 80.23 \\
            7                    & sherlyxxxxx    & 75.54 & 71.08 & 80.01 \\
            \rowcolor{gray!20}8  & xxxxl          & 75.53 & 71.00 & 80.06 \\
            9                    & aabbbcee       & 75.47 & 70.96 & 79.99 \\
            \rowcolor{gray!20}10 & skh            & 75.36 & 70.88 & 79.83 \\
            11                   & LJD            & 75.27 & 70.88 & 79.65 \\
            \rowcolor{gray!20}12 & dumplings      & 75.22 & 70.87 & 79.57 \\
            13                   & KirinCZW       & 74.95 & 70.55 & 79.34 \\
            \rowcolor{gray!20}14 & hkkk           & 74.47 & 70.11 & 78.82 \\
            15                   & hbx123573      & 73.83 & 69.44 & 78.22 \\
            \rowcolor{gray!20}16 & Tapallai       & 72.67 & 68.46 & 76.88 \\
            17                   & MahouShoujo    & 69.58 & 65.26 & 73.91 \\
            \rowcolor{gray!20}18 & j7991          & 57.35 & 52.72 & 61.97 \\
            19                   & jaspor         & 57.29 & 52.68 & 61.91 \\
            \thickhline
        \end{tabular}}%
    \label{tab:results_vos}%
    \vspace{-4mm}
\end{table}%

\section{VOS Track Teams and Methods}

The final performance of all teams in the VOS track is shown in \cref{tab:results_vos}. The top-performing teams are yahooo, yuanjie, and Sch89.89, achieving $\mathcal{J\&F}$ scores of 85.63, 85.26, and 80.76, respectively. The following sections provide detailed descriptions of the methods employed by the top four teams in the VOS track.

\subsection{PCL VisionLab team} \label{sec:pcl}
\teammembers{Deshui Miao\textsuperscript{1, 2}, Yameng Gu\textsuperscript{1},  Xin Li\textsuperscript{2}, Zhenyu He\textsuperscript{1,2}, Yaowei Wang\textsuperscript{2} and Ming-Hsuan Yang\textsuperscript{3}}
\teamaffiliations{\textsuperscript{1}Harbin Institute of Technology, Shenzhen \par 
\textsuperscript{2}Peng Cheng Laboratory \par  
\textsuperscript{3}University of California at Merced}

\input{sections/solutions/vos1/vos1_method}

\subsection{yuanjie team} \label{sec:yuanjie}
\teammembers{Jinming Chai, Qin Ma, Junpei Zhang, Licheng Jiao, Fang Liu}
\teamaffiliation{Intelligent Perception and Image Understanding Lab, Xidian University}

\input{sections/solutions/vos2/vos2_method}

\subsection{Xy-unu team} \label{sec:xy}
\teammembers{Xinyu Liu, Jing Zhang, Kexin Zhang, Xu Liu, Lingling Li}
\teamaffiliation{Intelligent Perception and Image Understanding Lab, Xidian University}

\input{sections/solutions/vos3/vos3_method}

\subsection{MVP-TIME team} \label{sec:time}
\teammembers{Feiyu Pan\textsuperscript{1}, Hao Fang\textsuperscript{2}, Runmin Cong\textsuperscript{2}, Wei Zhang\textsuperscript{2}, Xiankai Lu\textsuperscript{1}}
\teamaffiliations{
\textsuperscript{1}School of Software, Shandong University \par
\textsuperscript{2}School of Control Science and Engineering, Shandong University}

\input{sections/solutions/vos4/vos4_method}

\section{RVOS Track Teams and Methods}

The final performance of all teams in the RVOS track is shown in \cref{tab:results_rvos}. The top-performing teams are MVP-TIME, TXT, and CASIA\_IVA, achieving $\mathcal{J\&F}$ scores of 62.57, 60.40, and 60.36, respectively. The following sections provide the methods employed by the top three teams in the RVOS track.

\begin{table}[t]
    \renewcommand\arraystretch{1.2}
    \centering
    \caption{\textbf{RVOS Track results and final rankings.}}
    \vspace{-3mm}
    \setlength{\tabcolsep}{3.96mm}{\begin{tabular}{clrrr}
            \thickhline
             \rowcolor{gray!60} Rank & Team  & \multicolumn{1}{l}{$\mathcal{J}$} & \multicolumn{1}{l}{$\mathcal{F}$} & \multicolumn{1}{l}{$\mathcal{J\&F}$} \\
            \hline
            1                    & MVP-TIME       & 58.98 & 66.15 & 62.57 \\
            \rowcolor{gray!20}2  & TXT            & 57.02 & 63.78 & 60.40 \\
            3                    & CASIA\_IVA     & 56.88 & 63.85 & 60.36 \\
            \rowcolor{gray!20}4  & SaBoTaGe       & 56.89 & 63.83 & 60.36 \\
            5                    & BBBiiinnn      & 56.88 & 63.84 & 60.36 \\
            \rowcolor{gray!20}6  & bdc            & 56.82 & 63.74 & 60.28 \\
            7                    & BeverlyHam     & 56.78 & 63.54 & 60.16 \\
            \rowcolor{gray!20}8  & CHCH           & 56.69 & 63.42 & 60.06 \\
            9                    & PCL\_MDS       & 55.68 & 63.68 & 59.68 \\
            \rowcolor{gray!20}10 & nuk            & 55.37 & 62.08 & 58.73 \\
            11                   & Tapallai       & 54.22 & 60.71 & 57.46 \\
            \rowcolor{gray!20}12 & dgist\_lsh     & 51.72 & 61.21 & 56.47 \\
            13                   & forcom         & 51.52 & 58.99 & 55.26 \\
            \rowcolor{gray!20}14 & qian-long      & 50.27 & 59.42 & 54.85 \\
            15                   & neymarql       & 50.27 & 59.42 & 54.85 \\
            \rowcolor{gray!20}16 & liting         & 49.06 & 57.29 & 53.18 \\
            17                   & NanMu          & 47.98 & 56.80 & 52.39 \\
            \rowcolor{gray!20}18 & bai\_kai\_shui & 47.92 & 56.75 & 52.33 \\
            19                   & Jimmy46        & 47.44 & 51.79 & 49.62 \\
            \rowcolor{gray!20}20 & j7991          & 37.38 & 43.10 & 40.24 \\
            \thickhline
        \end{tabular}}%
    \label{tab:results_rvos}%
\end{table}%

\subsection{MVP-TIME team} \label{sec:time-r}
\teammembers{Hao Fang\textsuperscript{1}, Feiyu Pan\textsuperscript{2}, Xiankai Lu\textsuperscript{2}, Wei Zhang\textsuperscript{1}, Runmin Cong\textsuperscript{1}}
\teamaffiliations{\textsuperscript{1}School of Control Science and Engineering, Shandong University \par \textsuperscript{2}School of Software, Shandong University}

\input{sections/solutions/rvos1/rvos1_method}

\subsection{TXT team} \label{sec:txt}
\teammember{Tuyen Tran}
\teamaffiliation{Applied Artificial Intelligence Institute, Deakin University}

\input{sections/solutions/rvos2/rvos2_method}

\subsection{CASIA\_IVA team} \label{sec:casia}
\teammembers{Bin Cao\textsuperscript{1,2,3}, Yisi Zhang\textsuperscript{4}, Hanyi Wang\textsuperscript{2}, Xingjian He\textsuperscript{1}, Jing Liu\textsuperscript{1,2}}
\teamaffiliations{\textsuperscript{1}Institute of Automation, Chinese Academy of Sciences \par
\textsuperscript{2}School of Artificial Intelligence, University of Chinese Academy of Sciences \par
\textsuperscript{3}Beijing Academy of Artificial Intelligence\par
\textsuperscript{4}University of Science and Technology Beijing}

\input{sections/solutions/rvos3/rvos3_method}

\section{Conclusion \& Future Work}

This report presents a comprehensive overview of the methods and outcomes from the two tracks of the 6\nth LSVOS challenge. 
In the VOS track, the majority of approaches leverages memory networks to maintain long-term video context and improve object segmentation over extended sequences.
While In the RVOS track, there was an increased focus on integrating language models with temporal dynamics in videos, particularly building upon the MUTR framework, which highlights the understanding and processing the interplay between natural language and visual content over time. Also, it is noticeable that SAM-2 based methods are popular in both tracks  
However, despite these advancements, qualitative analysis reveals that accurately predicting object masks, especially in complex scenarios, remains a significant challenge.
We aim for the Large-scale Video Object Segmentation challenge to inspire and engage more researchers and participants in the challenging field of complex video object segmentation.

%% file: sections/solutions/vos1/vos1_method.tex
To address the challenges in video object segmentation (VOS), we present a robust method that incorporates semantic awareness and enhances query capabilities. Our approach introduces a novel fusion block that leverages both the semantic and detailed features derived from pretrained Vision Transformer (ViT) models. This strategy enables us to effectively manage complex variations in target appearance and resolve identification confusion among targets that look similar. Specifically, we integrate the CLS feature of the vision transformer with pyramid feature, enabling dense interatcion between frame regions and these multi-scale information for more refined detail integration. Moreover, we also design a discriminative query representation approach within the query transformer, which focuses on capturing the local features of the targets.
We describe key components as follows, and for more details please refer to \cite{li2024learningspatialsemanticfeaturesrobust}. 

\begin{figure}[ht]
\centering
\vspace{-3mm}
\includegraphics[width=0.99\textwidth]{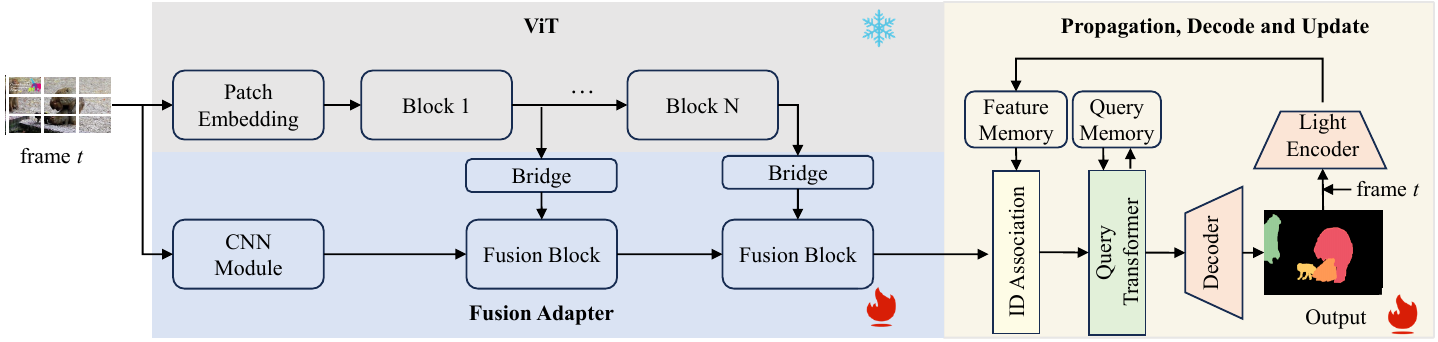}
\caption{Overall framework of PCL VisionLab team method, 1st place solution for 6th LSVOS Challenge in ECCV 2024.}
\vspace{-5mm}
\label{fig:overall}
\end{figure} 

\textbf{Spatial-Semantic Block. }
As illustrated in \cref{fig:overall}, the CLS response of ViT is fused with multi-scale features obtained from convolutional neural network architectures. Also, cross-attention is employed to facilitate semantic prior learning for VOS. Following this, multi-scale deformable attention is applied to understand the spatial relationships at different levels, aiding in the handling of complex shapes or disjoint parts.

\textbf{Discriminative Query Generation. }
We observed that it is not desirable to use the online prediction for building the Query Memory, as sometimes the artifacts from non-target areas are easily to be involved. This may diminish the discrimination capability of the target and cause accumulating erros as frame number goes.
In order to ensure effective propagation of target queries across frames, the target query memory is only updated using the most discriminating features of the target.

To achieve this, the similarities of the target query and each channel from the correlated fearure map are firstly evaluated, and only the most similar channel is extracted and used to update the target query.
We then use this identified feature from a new target sample to update the target queries. This is done by dynamically computing the interaction between the key query and significant pixel features in an additive manner. 

%% file: sections/solutions/vos2/vos2_method.tex
\begin{figure}[ht]
\centering
\vspace{-3mm}
\includegraphics[width=0.8\textwidth]{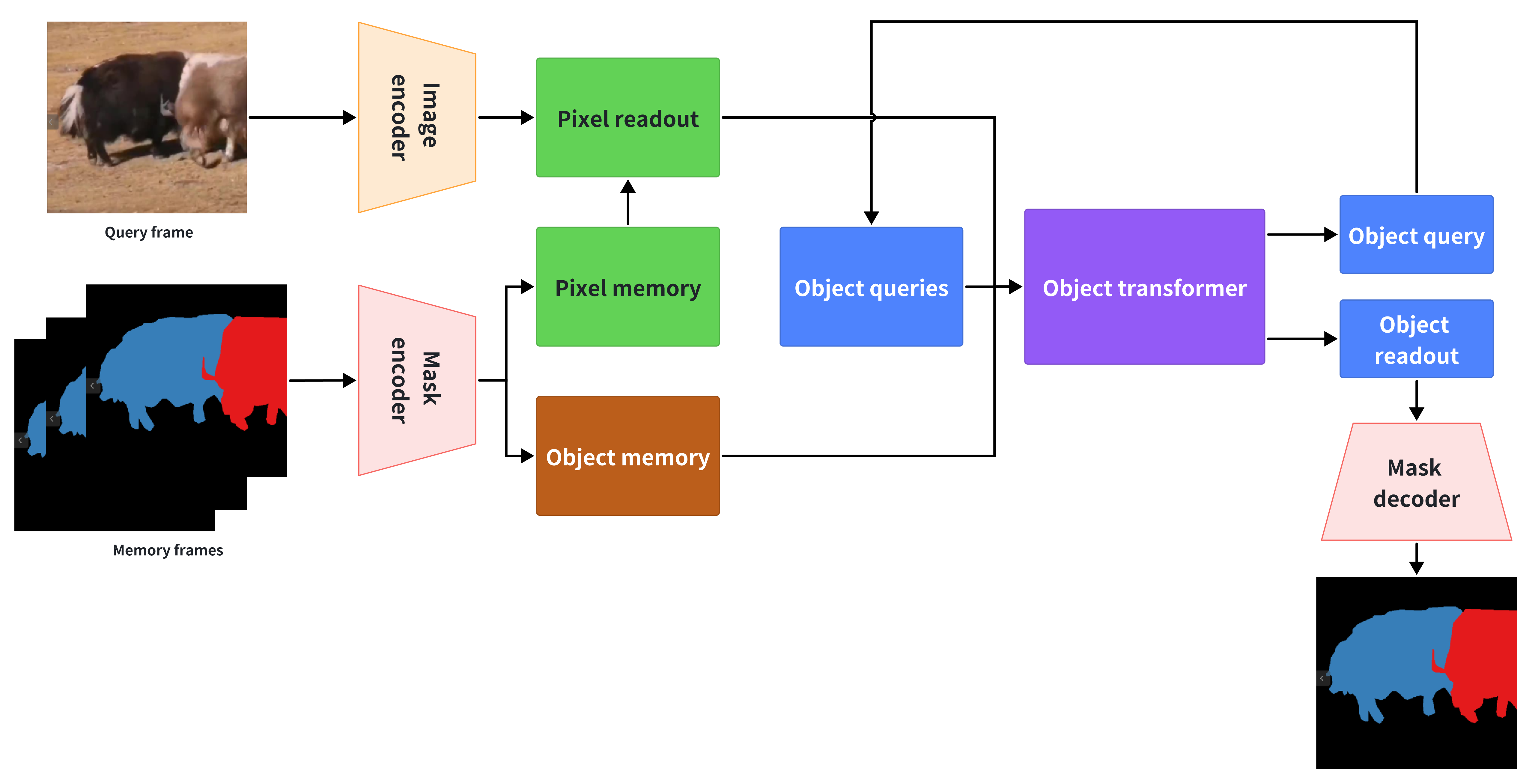} 
\caption{Workflow of the CSS-Segment. Image encoder is a streaming approach, consuming video frames as they become available. Mask encoder using convolutions and
summed element-wise with the image embedding. We store pixel memory and object memory representations from past segmented (memory) frames. Pixel memory is retrieved for the query frame as pixel readout, which bidirectionally interacts with object queries and object memory in the object transformer. The object transformer blocks enrich the pixel feature with object-level semantics and produce the final object readout for decoding into the output mask.} 
\label{fig:1} 
\vspace{-5mm}
\end{figure}

The proposed restoration framework contains four main steps, as shown in \cref{fig:1}: Image Encoder, Mask Encoder, Object Transformer, Object Memory.

\textbf{Image Encoder. }
The image encoder used in our framework is inspired by the design principles of SAM2 and is tailored for real-time processing of arbitrarily long videos. Unlike the ResNet50-based encoder used in the Cutie model, which may struggle with long sequences, our image encoder leverages a streaming approach. It processes video frames as they become available and is executed only once for the entire interaction. This design allows the encoder to provide unconditioned tokens (feature embeddings) that represent each frame effectively.
We utilize a hierarchical MAE with Hiera image encoder, which is specifically designed to handle multiscale features. This hierarchical structure enhances the encoder's ability to capture and represent complex, long-term video sequences more effectively than traditional models. By integrating multiscale features during decoding, our approach achieves superior data representation for long-sequence motion videos, addressing the limitations observed with the ResNet50-based encoder in the Cutie model.

\textbf{Mask Encoder. }
The design of our mask encoder is primarily inspired by the mask encoder used in SAM, offering a notable advancement over the ResNet18-based mask encoder utilized in the Cutie model. Our mask encoder integrates dense prompts (i.e., masks) with the image embeddings through a series of sophisticated convolutional operations. Specifically, masks are initially processed at a resolution 4 times lower than the input image, followed by further downscaling using two 2$\times$2 convolutions with stride 2, featuring output channels of 4 and 16 respectively. This is complemented by a final 1$\times$1 convolution that adjusts the channel dimension to 256. The entire process is enhanced with GELU activations and layer normalization at each stage.


\textbf{Object Transformer. }
\label{sec:object_transformer}
The Object Transformer, processes an initial readout $R_0 \in \mathbb{R}^{H \times W \times C}$, a set of $N$ end-to-end trained object queries $X \in \mathbb{R}^{N \times C}$, and object memory $S \in \mathbb{R}^{N \times C}$. It integrates these with $L$ transformer blocks to produce the final output. Here, $H$ and $W$ denote the image dimensions after encoding with a stride of 16. 
Before the first transformer block, the static object queries are summed with the dynamic object memory: $X_0 = X + S$. Each transformer block allows the object queries $X_{l-1}$ to attend to the readout $R_{l-1}$ bidirectionally, and vice versa, updating the queries to $X_l$ and the readout to $R_l$. The final readout $R_L$ of the last block is the output of the Object Transformer.

\textbf{Object Memory. }
The object memory, denoted as $S \in \mathbb{R}^{N \times C}$, stores a compact set of $N$ vectors that provide a high-level summary of the target object. This object memory is utilized in the aforementioned Object Transformer to offer target-specific features.
To compute $S$, we perform mask-pooling over all encoded object features. Specifically, given the object features $U \in \mathbb{R}^{THW \times C}$ and $N$ pooling masks $\{W_q \in [0, 1]^{THW} \mid 0 < q \leq N\}$, each mask $W_q$ is used to aggregate the features in $U$ into a summary vector for the object memory.
This pooling process ensures that the object memory captures relevant information from the encoded features, which is then leveraged for effective object representation in the transformer.

%% file: sections/solutions/vos3/vos3_method.tex
\begin{figure}[!htbp]
    \centering
    \vspace{-3mm}
    \includegraphics[width=\textwidth]{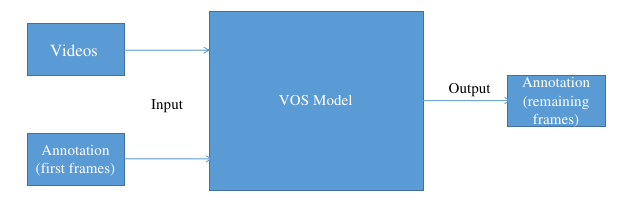}
    \caption{An overview of the Dual-Model VOS Enhancement VOS framework. The figure illustrates the key components of our approach, including the memory-based paradigm, pixel-level matching, and object query mechanism.}
    \label{fig:vos_framework}
    \vspace{-3mm}
\end{figure}

\begin{figure}[htbp]
    \vspace{-5mm}
    \centering
    \includegraphics[width=0.8\textwidth]{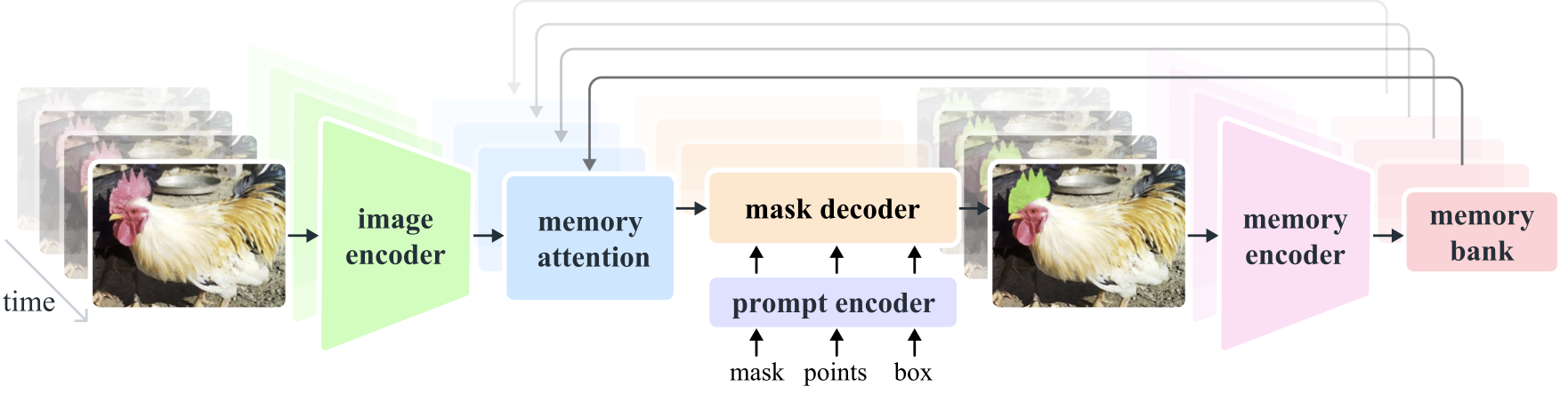}
    \caption{The SAM 2 architecture~\cite{ravi2024sam2}}
    \label{fig:sam2}
    \vspace{-5mm}
\end{figure}

Our approach is inspired by recent advancements in video object segmentation, specifically the SAM 2: Segment Anything in Images and Videos by Meta~\cite{ravi2024sam2} and the Cutie framework by Cheng \etal~\cite{cheng2023putting}.SAM2 is a unified model designed for both image and video segmentation, where an image is treated as a single-frame video. As shown in \cref{fig:sam2}, it generates segmentation masks for the object of interest, not only in single images but also consistently across video frames. A key feature of SAM2 is its memory module, which stores information about the object and past interactions. This memory allows SAM2 to generate and refine mask predictions throughout the video, leveraging the stored context from previously observed frames. The Cutie framework, on the other hand, operates in a semi-supervised video object segmentation (VOS) setting. It begins with a first-frame segmentation and then sequentially processes the following frames. Cutie is designed to handle challenging scenarios by combining high-level top-down queries with pixel-level bottom-up features, ensuring robust video object segmentation. Moreover, Cutie extends masked attention mechanisms to incorporate both foreground and background elements, enhancing feature richness and ensuring a clear semantic separation between the target object and distractors. Additionally, Cutie constructs a compact object memory that summarizes object features over the long term. During the querying process, this memory is retrieved as a target-specific object-level representation, which aids in maintaining segmentation accuracy across the video.

%% file: sections/solutions/vos4/vos4_method.tex
As shown in \cref{fig:sam2}, SAM 2~\cite{ravi2024sam2} supports point, box, and mask prompts on individual frames to define the spatial extent of the object to be segmented across the video. For image input, the model behaves similarly to SAM. A promptable and light-weight mask decoder accepts a frame embedding and prompts on the current frame and outputs a segmentation mask for the frame. Prompts can be iteratively added on a frame in order to refine the masks. Unlike SAM, the frame embedding used by the SAM 2 decoder is not directly from an image encoder and is instead conditioned on memories of past predictions and prompted frames. It is possible for prompted frames to also come “from the future” relative to the current frame. Memories of frames are created by the memory encoder based on the current prediction and placed in a memory bank for use in subsequent frames. The memory attention operation takes the per-frame embedding from the image encoder and conditions it on the memory bank to produce an embedding that is then passed to the mask decoder.


%% file: sections/solutions/rvos1/rvos1_method.tex
\begin{figure}[t]
\begin{center}
    \vspace{-3mm}
\includegraphics[width=1\linewidth]{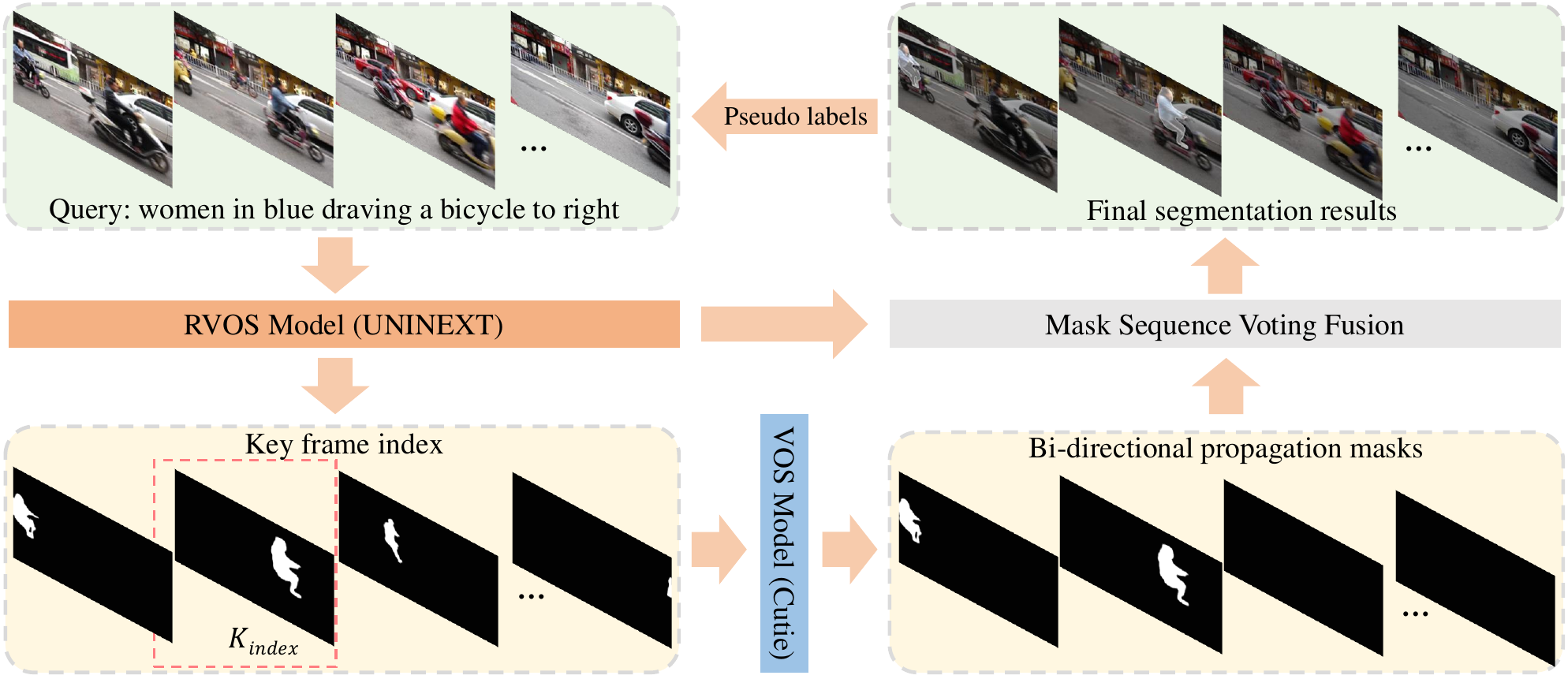}
\end{center}
\caption{The overview architecture of the proposed method from MVP-TIME: The 1st Solution for LSVOS Challenge RVOS Track.}
\vspace{-3mm}
\label{fig:rvos1_model}
\end{figure}

\label{sec:metho}
The input of RVOS contains a video sequence $\mathcal{V} = \left\{v_t\in \mathbb{R}^{3 \times H \times W} \right\}_{t=1}^T $ with \textit{T} frames and a corresponding referring expression $\mathcal{E} = \left\{ e_l \right\}_{l=1}^L $ with \textit{L} words. Our solution consists of three steps: Backbone, Post-process, and Semi-supervised. The overall architecture of the proposed method is illustrated in ~\cref{fig:rvos1_model}. 

\textbf{Backbone}
\label{sec:backbone}
We adopt the state-of-the-art RVOS model UNINEXT~\cite{yan2023universal} as our backbone to obtain mask sequences $\mathcal{S} = \{s_t\}_{t=1}^T$ that are correlated with language descriptions.
\begin{equation}
    \mathcal{S} = \mathcal{F}^{rvos}\left( \mathcal{V}, \mathcal{E}\right),
\end{equation}
where $\mathcal{F}^{rvos}$ denotes the UNINEXT model. UNINEXT reformulates diverse instance perception tasks into a unified object discovery and retrieval paradigm, and achieved surprising performance after joint training on multiple datasets. So we fine-tuned the official pre-training weights provided on MeViS.

\textbf{Post-process}
\label{sec:post-process}
Previous challenge solutions~\cite{luo20241st,hu20221st} have shown that using a semi-supervised VOS algorithm can further improve the accuracy of segmentation results. The general procedure are first selecting the key-frame index of mask sequences probability $\mathcal{P}$ from RVOS model, then using VOS model to perform forward and backward propagation. It can be formulated as:
\begin{equation}
\begin{gathered}
    \mathcal{K}_{index} = argmax(\mathcal{P}), \\
    \mathcal{M} = \left[\mathcal{F}^{vos}\left(\{s_{i}\}_{i=K_{index}}^{0}\right), \mathcal{F}^{vos}\left(\{s_{j}\}_{j=K_{index}}^{T}\right)\right], \\
\end{gathered}
\end{equation}
where $\mathcal{P} = \left\{ p_k \in \mathbb{R}^1 \right\}_{k=1}^T$, $\mathcal{F}^{vos}$ denotes the VOS model for post-process. We adopt the state-of-the-art VOS model Cutie~\cite{cheng2023putting} for post-process.

In our experiment, we find that post-process dose not improve the mask quality of all videos. The reason is that MeViS is a multi-object dataset, and the mask with the highest probability output by UNINEXT may not necessarily include all specified objects. This may not be a problem with UNINEXT, it could just be that only a single object appeared in that frame. Therefore, we select the $N$ masks with the highest probability in the RVOS model for VOS inference and fuse them with the mask sequence output by the original RVOS model.
\begin{equation}
    \mathcal{M} = \mathcal{F}^{fuse}\left( \mathcal{S}, \mathcal{M}^{N}\right),
\end{equation}
where $\mathcal{M}^{N}$ is the $N$ sets of mask sequences output by Cutie, $\mathcal{F}^{fuse}$ denotes pixel-level binary mask voting. If there are more than $(N+1)/2$ pixels with a value equal to 1, we divide the pixel into the foreground, otherwise, it is divided into the background.

\textbf{Semi-supervised}
\label{sec:semi-supervised}
The post-processing result $\mathcal{M}$ is significantly better than the backbone result $\mathcal{S}$, thus the predicted results on the validation set of MeViS dataset can be served as pseudo ground truth object masks of validation set. We then re-finetune the backbone model UNINEXT on validation set with pseudo labels. This semi-supervised approach~\cite{cao2022second} is also employed on the testing set. Finally, performing further post-processing after fine-tuning can further improve performance.

%% file: sections/solutions/rvos2/rvos2_method.tex
\begin{figure}
\begin{centering}
    \vspace{-3mm}
\includegraphics[width=0.99\textwidth]{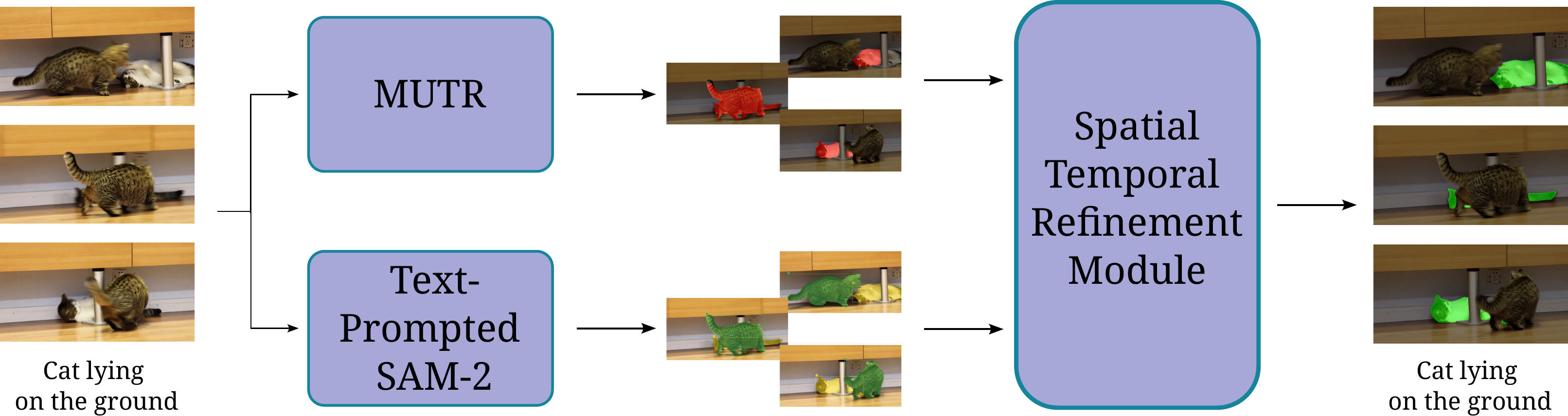}
\par\end{centering}
\caption{We first extract the main noun from the given textual query (\eg,
``Cat'') and use it as input for the \emph{Text-Prompted
SAM-2}. This module essentially combines Grounding Dino and SAMv2.
Grounding Dino detects all bounding boxes of instances belonging to
the specified object category. These boxes are then used as prompt
input for the SAMv2 model, resulting in a sequence of spatio-temporal
masks. Concurrently, a fine-tuned MUTR model is employed to generate
coarse masks from the input video. These initial masks are then subjected
to refinement within the Spatial-Temporal Refinement Module, resulting
in final segmentation masks with improved temporal consistency.\protect\label{fig:rvos2_overal}}
\vspace{-3mm}
\end{figure}

The overall proposed framework is presented in \cref{fig:rvos2_overal}.
We initially employ SAM-2 to extract spatio-temporal masks containing
tracking-related details. Simultaneously, we fine-tune the MUTR model
on MeViS to generate initial coarse spatio-temporal masks based on
the given video and textual description. We define these coarse masks
as $M_{c}=\left\{ u^{t}\right\} _{t=1}^{T},$ where $T$ is the number
of frame in video. The resulting raw masks undergo further refinement
in the Spatial-Temporal Refinement Module to yield the
final segmentation mask with enhanced temporal consistency.

\textbf{Video object tracking with textual prompt. }
Since the original SAM-2 requires initial inputs of either points,
boxes, or masks to track visual objects within a video, additional
processing steps are necessary to construct a video object tracking
system using SAM-2 with textual prompt input. First, given a descriptive
sentence, we employ a language processing tool Berkeley Neural Parser
\cite{kitaev} to extract the main noun (\eg, ``Cat''
in \cref{fig:rvos2_overal}), which designates the target object category
for tracking. Subsequently, we utilize the open-vocabulary object
detection model, Grounding DINO \cite{liu2023grounding}, to extract
bounding boxes encompassing the target object. These bounding boxes
serve as input prompts for the SAM-2 model. SAM-2 produces a set of
spatio-temporal masks, termed `masklets'. The number of masklets corresponds
to the quantity of distinct instances detected for the given object
category. Formally, we denote the tracking results from SAM-2 as $M_{t}=\left\{ v_{i}^{t}\right\} $,
where $i$ ranges from $1$ to $N$ and $t$ ranges from $1$ to $T$.
Here, $N$ is the number of instances detected for the given
object category, and $T$ denotes the number of frames in the input
video.

\textbf{Spatial-Temporal Refinement for Consistent Semantic Segmentation. }
The pseudo code is outlined in \cref{alg:psedu-code}. We
first divide the entire video, which consists of $T$ frames, into
non-overlapping sequences with a window size of $W$. The proposed
module, which takes input as coarse masks $M_{c}$ and tracked masks
$M_{t}$, is executed on each sequence individually. With slight abuse
of notation, we also will use $M_{t}=\left\{ v_{i}^{t}\right\} $
with $i$ ranges from $1$ to $W$ to denote the tracking results
of instance $v_{i}$ within the window size $W$. At each time step,
we calculate the Fraction of Overlap $f_{i}^{t}$ between each tracked
instance $v_{i}^{t}$ and the coarse segmentation mask $u^{t}$:

\begin{equation}
f_{i}^{t}=\frac{\text{Intersection\ensuremath{\left(v_{i}^{t},u^{t}\right)}}}{\text{Area(\ensuremath{v_{i}^{t}})}}.
\end{equation}
$f_{i}^{t}$ is calculated as the ratio of the intersection area between
the tracked instance $v_{i}^{t}$ and the coarse mask $u^{t}$ to
the total area of instance $v_{i}^{t}$. The Fraction of Overlap $f_{i}^{t}$
indicates the proportion of instance $i$ at time step $t$ that overlaps
with the coarse mask predicted by the MUTR model. If $f_{i}^{t}$
exceeds a threshold $\tau$, we infer that instance $i$ is present
at time step $t$ and add its index to the component list $C_{t}$.
As a result, $C_{t}$ represents the combination of components at
the time step t. This process is repeated across all time steps within
the window size $W$, yielding the set $\mathcal{C}=\left\{ C_{t}\right\} ,$
where each element $C_{t}$ captures the specific combination of components
at its respective time step $t$. We expect that within a given window
size, the spatio-temporal masks should remain consistent. Therefore,
we select $C_{\text{sel}}$ as the combination of components that
appears most frequently in the set $\mathcal{C}$ for refinement.
The refined spatio-temporal masks $M_{r}=\left\{ m^{t}\right\} $
are derived by composing all instances listed in $C_{\text{sel}}$.
If $C_{\text{sel}}=\emptyset$, meaning that the predicted instances
from the MUTR model are not included in the tracking output, we retain
the original prediction without refinement. 

\SetKwComment{Comment}{/* }{ */} 
\vspace{-3mm}
\begin{algorithm*}
    \footnotesize
\KwIn{ Coarse segmentation masks $M_{c}=\left\{ u^{t}\right\} _{t=1}^{W},$; Tracked masks $M_{t}=\left\{ v_{i}^{t}\right\} _{i=1,t=1}^{N,W}$
}
\KwOut{Refined mask $M_{r}=\left\{ m^{t}\right\} _{t=1}^{W}$ }
\hrulealg
\For{$t=1$ to $W$} {

\ForEach{$v_{i}^{t}$}{

Calculate $f_{i}^{t}$ \tcp*[f]{Refer to equation 1}\;

\lIf{$f_{i}^{t}>\tau$}{Add $i$ to $C_{t}$ }

} }

Set $C_{\text{sel}}$ as the combination that appears most frequently
in $\left\{ C_{t}\right\} $\; 

Obtain the refined mask $M_{r}=\left\{ m^{t}\right\} _{t=1}^{W}$
by composing all instances included in $C_{\text{sel}}$\;
\caption{Spatial-Temporal Refinement for Consistent Segmentation.\protect\label{alg:psedu-code}}
\end{algorithm*}
\vspace{-12mm}

%% file: sections/solutions/rvos3/rvos3_method.tex
\begin{figure*}[ht]
  \centering
  \vspace{-3mm}
   \includegraphics[width=1.0\linewidth]{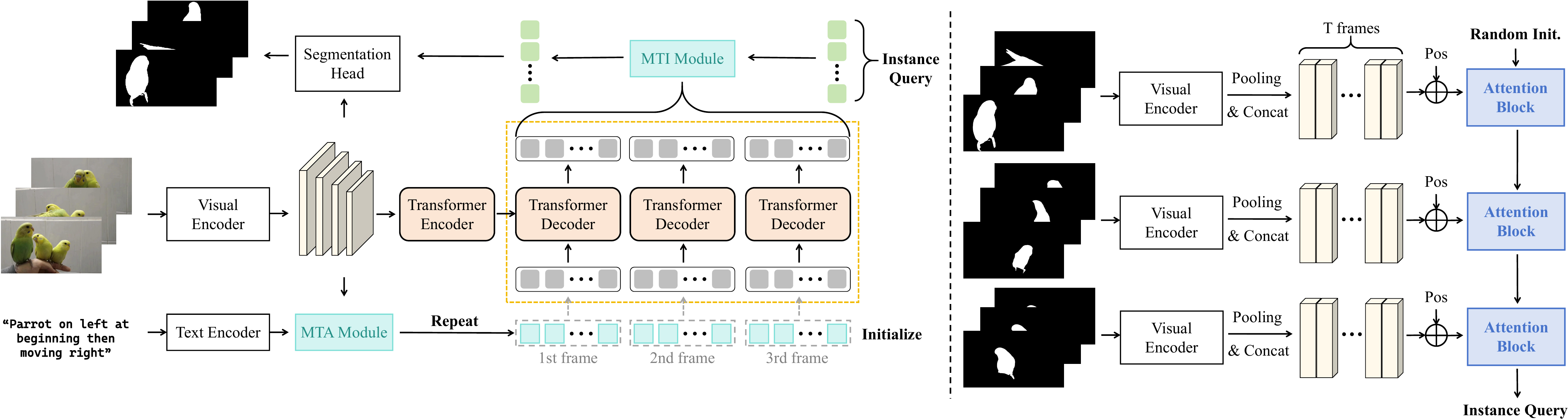}
   \vspace{-3mm}
   \caption{\textbf{The architecture of CASIA IVA team method.} We employ MUTR as our basic model (\textbf{Left}). We introduce instance masks and employ an attention block and a sequential mechanism to aggregate instance information into a query (\textbf{Right}).}
   \vspace{-3mm}
   \label{fig:rvos3_model_architecture}
\end{figure*}

\textbf{Overview.}
Our solution contains three components: MUTR-based model; instance retrieval model and fusion strategy. The architecture of MUTR-based solution is shown in \cref{fig:rvos3_model_architecture}. To improve the consistency of results, we introduce proposal instance masks into MUTR for query initialization. After prediction, we employ HQ-SAM to refine prediction masks by sampling key points as prompts. 

\textbf{MUTR-based Model. }
MUTR (\textbf{M}ultimodal \textbf{U}nified \textbf{T}emporal transformer for \textbf{R}eferring video object segmentation) was proposed in \cite{yan2024referred} and has shown superior performance on Ref-Youtube-VOS and Ref-DAVIS17. MUTR adopts a DETR-like style model. Compared with other methods, MUTR introduces two core modules, i.e. MTI (\textbf{M}ulti-object \textbf{T}emporal \textbf{I}nteraction module), MTA (\textbf{M}ulti-scale \textbf{T}emporal \textbf{A}ggregation module).  


Specifically, we attempt to introduce instance masks to initialize the video-wise query $\mathcal{Q}$ in MTI decoder. Thanks to the superior performance of DVIS on VIS, we employ DVIS for mask generation, which extracts all instance masks in a video clip as follows:
\begin{equation}
    m_{i} = \text{DVIS}(\text{I}), \ m_{i} \in \mathbb{R}^{\ T \times H \times W}
\end{equation}
where $\text{I} \in \mathbb{R}^{\ T \times H \times W \times \text{3}}$ is the input video clip, $m = \{m_{i}\}^{K}_{i=1}$ denotes the set of instance masks, $K$ is the number of instances in a video clip and $T$ is the number of frames.

The motion property is a significant aspect that can distinguish different objects. Therefore, we inject motion cues into instance features. Given a multi-frame instance binary mask $m_{i}$, we calculate the bounding box of this object for each frame and obtain the positional information as follows:
\begin{equation}
   p_{i,t} = (x^{i,t}_{min}, y^{i,t}_{min}, x^{i,t}_{max}, y^{i,t}_{max}, x^{i,t}_{c}, y^{i,t}_{c}, w_{i,t}, h_{i,t})
\end{equation}
where $(x^{i,t}_{min}, y^{i,t}_{min}), (x^{i,t}_{max}, y^{i,t}_{max}), (x^{i,t}_{c}, y^{i,t}_{c}), w_{i,t}$ $h_{i,t}$ are normalized top-left coordinates, bottom-right coordinates, center coordinates, width and height of bounding box respectively, $t$ is the index of video frames.

Next, we utilize a visual encoder to extract multi-scale visual features of instance masks and inject the instance trajectory into visual features as follows:
\begin{equation}
    \mathcal{F}_{i,j} = \text{Visual\_Backbone}(m_{i}) + W(p_{i}), \ \mathcal{F}_{i,j} \in \mathbb{R}^{\ T \times h_{j} \times w_{j} \times c_{j} }
\end{equation}
where $c_{j}$ is the channel of $j$ level visual feature and $W$ is a linear layer. After feature extraction, we utilize a projection layer on multi-scale visual features to align dimension with video features and perform average pooling along spatial dimension to obtain instance features as follows: 
\begin{equation}
    \mathcal{F}^{'}_{i,j} = \text{Pooling}(\text{Proj}(\mathcal{F}_{i,j})), \  \mathcal{F}_{i,j} \in \mathbb{R}^{\ T \times C }
\end{equation}

For simplicity, we only explain our solution utilizing the single-level visual feature. To aggregate all instance information into an instance query, we design an attention block and adapt sequential mechanisms as follows: 
\begin{equation}
    \mathcal{Q}_{i} = \text{Block}(\mathcal{Q}_{i-1},\mathcal{F}^{'}_{i}), \ 1 \leq i \leq K 
\end{equation}
where ${Q}_{i} \in \mathbb{R}^{\ N \times C}$ is the instance query and $N$ is the number of queries. ${Q}_{0}$ is randomly initialized. The designed attention block consists of a cross-attention layer, a set of self-attention layers, and FFN layers. After that, we utilize this query with instance information to replace the randomly initialized video-wise query fed to MTI decoder.

\textbf{HQ-SAM for Spatial Refinement.}
We adopt HQ-SAM \cite{ke2024segment,ke2022video} with ViT-L as our mask refiner. Given the predicted result from MUTR of each clip, we first determine the coordinates of the bounding box by selecting the maximum and minimum horizontal and vertical coordinates of the points along the boundary of the mask. Next, we uniformly sample 10 coordinates within the predicted mask as positive points and 5 coordinates out of the mask but within the bounding box as negative points. The sampled points are then fed into the mask decoder of HQ-SAM as prompts to generate the refined masks.

\textbf{Instance Retrieval Model.}
We employ a classification model which predict the valid masks sequence under the language expression from candidates generated by VIS model. Specifically, we choose DVIS \cite{zhang2023dvis} to generate the candidate masks with long frame length. The classification follows a simple architecture with Swin-Large and RoBERTa serving as vision and language backbone, respectively. The corresponding vision features are fed into a standard cross-attention module as query with language features as key and value. 
The obtained features are consequently averaging pooled at the candidate mask level, following a one-hot classifier to obtain the valid mask sequence result under present language expression. 

\textbf{Fusion Strategy.}
We design a fusion strategy to fuse predicted results from two models both frame-level and instance-level. First, we filter results of MUTR-based model with noise and retrieve instance from 
results of instance retrieval model utilizing IOU in a frame-independent manner. Then, we utilize the frame-level fusion results to retrieve the instance from the whole video utilizing IOU.